\documentclass[dvipsnames,format=sigconf,nonacm]{acmart}

\usepackage{float}
\usepackage{comment}
\usepackage{amsmath}
\usepackage{tabularx}
\usepackage{makecell}
\usepackage{multirow}
\usepackage{enumitem}
\usepackage{siunitx}
\usepackage{mathtools}
\usepackage{svg}[inkscapelatex=false]
\usepackage{graphicx}
\usepackage{url}
\usepackage{tikz}
\usepackage[edges]{forest}
\usepackage[font=small,labelfont=bf,belowskip=5pt]{caption}
\usepackage{subcaption}

\def\vec#1{\mathchoice{\mbox{\boldmath$\displaystyle#1$}}
{\mbox{\boldmath$\textstyle#1$}}
{\mbox{\boldmath$\scriptstyle#1$}}
{\mbox{\boldmath$\scriptscriptstyle#1$}}}

\DeclarePairedDelimiter\floor{\lfloor}{\rfloor}
\AtBeginDocument{%
  }

\copyrightyear{2025}
\acmYear{2025}
\setcopyright{cc}
\setcctype{by-nc-sa}
\acmConference[GECCO '25]{Genetic and Evolutionary Computation
Conference}{July 14--18, 2025}{Malaga, Spain}
\acmBooktitle{GECCO '25,
July 14--18, 2025, Malaga, Spain}\acmDOI{10.1145/3712256.3726418}
\acmISBN{979-8-4007-1465-8/2025/07}

\begin{document}

\title[Incremental Distribution Estimation in RV-GOMEA]{More Efficient Real-Valued Gray-Box Optimization through Incremental Distribution Estimation in RV-GOMEA}

\author{Renzo J. Scholman}
\email{renzo.scholman@cwi.nl}
\orcid{0000-0003-2813-015X}
\affiliation{%
  \institution{Centrum Wiskunde \& Informatica}
  \city{Amsterdam}
  \country{The Netherlands}
}
\affiliation{%
  \institution{Delft University of Technology}
  \city{Delft}
  \country{The Netherlands}
}

\author{Tanja Alderliesten}
\email{t.alderliesten@lumc.nl}
\orcid{0000-0003-4261-7511}
\affiliation{%
  \institution{Leiden University Medical Center}
  \city{Leiden}
  \country{The Netherlands}
}

\author{Peter A.N. Bosman}
\email{peter.bosman@cwi.nl}
\orcid{0000-0002-4186-6666}
\affiliation{%
  \institution{Centrum Wiskunde \& Informatica}
  \city{Amsterdam}
  \country{The Netherlands}
}
\affiliation{%
  \institution{Delft University of Technology}
  \city{Delft}
  \country{The Netherlands}
}

\renewcommand{\shortauthors}{Scholman et al.}

\begin{abstract}
The Gene-pool Optimal Mixing EA (GOMEA) family of EAs offers a specific means to exploit problem-specific knowledge through linkage learning, i.e., inter-variable dependency detection, expressed using subsets of variables, that should undergo joint variation. Such knowledge can be exploited if faster fitness evaluations are possible when only a few variables are changed in a solution, enabling large speed-ups. The recent-most version of Real-Valued GOMEA (RV-GOMEA) can learn a conditional linkage model during optimization using fitness-based linkage learning, enabling fine-grained dependency exploitation in learning and sampling a Gaussian distribution. However, while the most efficient Gaussian-based EAs, like NES and CMA-ES, employ incremental learning of the Gaussian distribution rather than performing full re-estimation every generation, the recent-most RV-GOMEA version does not employ such incremental learning. In this paper, we therefore study whether incremental distribution estimation can lead to efficiency enhancements of RV-GOMEA. We consider various benchmark problems with varying degrees of overlapping dependencies. We find that, compared to RV-GOMEA and VKD-CMA-ES, the required number of evaluations to reach high-quality solutions can be reduced by a factor of up to 1.5 if population sizes are tuned problem-specifically, while a reduction by a factor of 2-3 can be achieved with generic population-sizing guidelines.
\end{abstract}




\maketitle
\section{Introduction}
Taking a Gray-Box Optimization (GBO) approach to exploit problem-specific knowledge, can substantially enhance the performance of EAs. Examples include optimization of a billion-parameter problem using problem-specific recombination operators~\cite{deb_breaking_2016}, adapted crossover operators for satellite scheduling~\cite{whitley_scheduling_2023}, and tailored mutation operators for wireless communication networks~\cite{neumann_diversity_2023}.

Sometimes, a problem allows for efficient partial evaluations, meaning that objective values can be efficiently recalculated when only a few variables in a solution are modified, greatly reducing total optimization time, see, e.g.,~\cite{bouter_achieving_2021,andreadis_morea_2023, luong_application_2018}.

EAs in the family of Gene-pool Optimal Mixing EAs (GOMEA) are especially suited for exploiting partial evaluations due to the use of optimal mixing~\cite{thierens_optimal_2011}. In GOMEA, subsets of interdependent variables are leveraged, called linkage sets, in the GOM variation operator. By only changing the variables in a linkage set, and keeping the change only if the solution did not get worse, partial evaluations can be effectively exploited, which leads to performance benefits if the linkage sets are well-aligned with problem structure.

State-of-the-art performance can be achieved on real-valued problems using the Real-Valued GOMEA (RV-GOMEA), especially in gray-box scenarios, in which Black-Box Optimization (BBO) approaches can be vastly outperformed~\cite{bouter_achieving_2021}.

Recently, a new version of RV-GOMEA was introduced that learns conditional linkage sets using fitness-based linkage learning techniques during optimization~\cite{andreadis_fitness-based_2024}. The use of such linkage models in RV-GOMEA has been essential to also have an performance advantage over black-box methods on optimization problems with overlapping dependencies. The Gaussian distributions pertaining to each linkage set are however still estimated in each generation using only the best-performing solutions in the current population. Doing so discards information from the previous generation and, therefore, typically results in requiring larger population sizes to make reliable estimations of covariances.

The Gaussian sampling model in RV-GOMEA is based on AMaLGaM, a real-valued EDA~\cite{bouter_achieving_2021}. It is known that the Gaussian in AMaLGaM can be learned incrementally across generations~\cite{bosman_empirical_2009}. This can typically reduce the required population size and often, the required number of function evaluations.

Similar incremental mechanisms are included in the state-of-the-art real-valued EAs CMA-ES~\cite{hansen_completely_2001} and NES~\cite{wierstra_natural_2008}.
A version of RV-GOMEA, called RV-GOMEA$_{C}$, exists that uses the CMA-ES sampling model, but for decomposable problems with moderately sized subfunctions, performance differences with RV-GOMEA are small, with RV-GOMEA sometimes outperforming RV-GOMEA$_{C}$~\cite{bouter_achieving_2021}.

Still, the parameter settings of the CMA-ES sampling model were not reconsidered in the integration with RV-GOMEA. Proper calibration of incremental learning may however still lead to achieving even better results, since the incremental AMaLGaM was established entirely through empirical optimization~\cite{bosman_empirical_2009}. In this paper, we therefore reconsider the incremental learning model for AMaLGaM, but then within the recent-most version of RV-GOMEA, and revisit optimizing the parameters of this model.
We refer to the resulting new EA as the incremental RV-GOMEA (iRV-GOMEA).

In the remainder of this paper, we first explain how incremental learning is incorporated into RV-GOMEA in Section~\ref{SectioniRVGOMEA}. The same section describes how we tune iRV-GOMEA on various problems and determine population-sizing guidelines. Then, in Section~\ref{SectionExperiments}, we describe the setup of our experiments, testing the resulting iRV-GOMEA in various ways and comparing its performance with that of RV-GOMEA, RV-GOMEA$_{C}$, and VkD-CMA-ES~\cite{akimoto_online_2016,akimoto_projection-based_2016}. The results of the experiments are presented in Section~\ref{SectionResults}. We present a discussion in Section~\ref{SectionDiscussion} and our final conclusions in Section~\ref{SectionConclusion}.

\color{black}
\section{Background}\label{SectionBackground}
\subsection{Incremental Distribution Estimation}
Incremental learning typically takes the following form:
\begin{equation}
\vec{\theta}(t) = (1 - \eta)\vec{\theta}(t - 1) + \eta \hat{\vec{\theta}}(t).
\label{eq:memory}
\end{equation}

In our setting, $\hat{\vec{\theta}}(t)$ are the probability distribution parameters as estimated from the population in generation $t$, and $\vec{\theta}(t)$ are the incrementally learned parameters with a learning rate $\eta \in [0,1]$.

A general function class has been derived in literature to estimate an effective learning rate for distribution parameters in an EDA in which truncation selection is used and the distribution parameters are estimated from the selected solutions~\cite{bosman_empirical_2009}.
The function class is defined using the selection size $|\mathbf{\mathcal S}|$, problem dimensionality $\ell$, and three function-class parameters $\vec{\alpha}=(\alpha_0, \alpha_1, \alpha_2)$ as follows:
\begin{equation}
\eta = f^{\eta}_{\vec{\alpha}}(|\mathbf{\mathcal S}|, \ell) = 1 - \exp \left( \dfrac{ \alpha_0 |\mathbf{\mathcal S}|^{\alpha_1} }{\ell^{\alpha_2}} \right).
\label{eq:learning_rate_function}
\end{equation}

This function class can be instantiated by fitting it to a dataset of $N$ samples $\{(\eta^i, (|\mathbf{\mathcal S}|^i, \ell^i))\}, i \in \{0, 1, \ldots , N-1\}$, i.e., by performing non-linear regression with a typical sum-of-squares loss function:
\begin{equation}
\min_{\vec{\alpha}} \sum_{i=0}^{N-1} \left( \eta^i - f^{\eta}_{\vec{\alpha}}(|\mathbf{\mathcal S}|^i, \ell^i) \right)^2.
\label{eq:learning_rate_fitting}
\end{equation}

Performing this non-linear optimization task can be done using, e.g., an EA.
To regress the $\alpha_{i}$ parameters, a dataset is needed consisting of the (near-)best values for the learning rate associated with various selection sizes and problem dimensionalities.
This can be obtained through experimentation, i.e., by trying various values for, or optimizing, the learning rate for different population sizes and different problem dimensionalities.
Such experimentation can be done for a \emph{set} of problems, resulting in a more general instantiation of the function class that is well-suited for a class of problems.

\subsection{Gray-box Optimization}
Consider a problem where the objective function $f(\mathbf{x}) : \mathbb{R}^\ell \rightarrow \mathbb{R}$ is to be minimized and a solution is denoted by $\mathbf{x} = (x_0,\dots,x_{\ell-1})$.

An objective function $f(\mathbf{x})$ permitting partial evaluations, is composed of $q$ non-decomposable sub-functions $f_0,\dots,f_{q-1}$.
Now, let $\mathcal{I} = \left\{\mathcal{I}_0,\dots,\mathcal{I}_{q-1}\right\}$, $\mathcal{I}_{i} \subseteq \{0,1,\ldots,\ell-1\}$ denote the subsets of indices that are involved with the $q$ sub-functions.
The variables pertaining to the $i$--th subset and subfunction can be denoted as $\mathbf{x}_{\mathcal{I}_i}$. After a modification is made (i.e., during variation) to one or more variables $x_j$, all sub-functions $f_i(\mathbf{x}_{\mathcal{I}_i})$ are evaluated for which $j \in \mathcal{I}_i$ and subsequently, the fitness of the solution can be recomputed from the values of the subfunctions $f_i$. For more details on this formulation and its use to perform partial evaluations, see, e.g.,~\cite{bouter_achieving_2021}.

In this work, we consider the cost of the partial evaluation of $f_i$ as $|\mathcal{I}_i|$ / $|\mathcal{I}|$.
The cost of changing one or more variables $x_j$ is then computed as the accumulated cost of each affected sub-function $f_i$.

Two variables $x_u$ and $x_v$ are considered dependent if there exists a sub-function $f_i$ for which $\{u,v\} \subseteq \mathcal{I}_i$ holds.
A Variable Interaction Graph (VIG) is an undirected graph $G = (V,E)$ where
each vertex $v \in V$ corresponds to a variable $x_v$ and for each pair of dependent variables there exists an edge $(u,v) \in E$, see, e.g.,~\cite{whitley_gray_2016, tinos_partition_2021}.

\subsection{RV-GOMEA}
RV-GOMEA~\cite{bouter_achieving_2021} is a variant of GOMEA for solutions with real-valued variables. In RV-GOMEA, a sampling model is used to generate new solutions that comes from the EDA called AMaLGaM~\cite{bosman_enhancing_2008}.
From the population $\mathbf{\mathcal P}$, AMaLGaM selects the set $\mathbf{\mathcal S}$ of $\lfloor \tau|\mathbf{\mathcal P}|\rfloor$, $\tau \in [1/|\mathbf{\mathcal P}|,1]$ best-performing individuals and estimates a Gaussian (normal) distribution $P^{\mathcal N}(\cdot)$ from $\mathbf{\mathcal S}$.
Subsequently, variation is performed by sampling new solutions from $P^{\mathcal N}(\cdot)$.

RV-GOMEA uses the Family Of Subsets (FOS) linkage model $\mathbf{\mathcal{F}}=\{\mathbf{F}^{0},\mathbf{F}^{1},\ldots,\mathbf{F}^{|\mathbf{\mathcal{F}}|-1}\}$, which comprises a set of FOS elements $\mathbf{F}^{i}$, also called linkage sets. Each linkage set is a set of variable indices $\mathbf{F}^{i} \subseteq \{0,1,\ldots,\ell-1\}$ that identify variables that are considered to be interdependent and must be processed \emph{jointly} during variation.
For each linkage set $\textbf{F}^{i}$, RV-GOMEA estimates a multivariate Gaussian distribution $P^{\mathcal N,i}(\cdot)$ using the same procedure as in AMaLGaM.

During variation with the GOM operator, all linkage sets $\textbf{F}^i$ are considered in a random order and for each solution in the population, new values are jointly sampled from $P^{\mathcal N,i}(\cdot)$ for each of the variables in the linkage set. Changes are in principle only kept if the solution has not become worse. However, in the original, non-conditional, formulation of RV-GOMEA, with a probability of 5\%, changes were always accepted~\cite{bouter_achieving_2021}.

As in AMaLGaM~\cite{bosman_enhancing_2008}, the techniques known as Anticipated Mean Shift (AMS) and Adaptive Variance Scaling (AVS) are used in RV-GOMEA.
Through AMS, a part of the population is moved in the direction in which the estimated distribution mean shifted in the recent-most subsequent generations, enhancing optimization on slope-like regions in the search space.
AVS is used to counteract the variance-diminishing effect of selection whenever improvements are found far away from the mean. This is done through the use of distribution multipliers $c^{mult}_{\textbf{F}^i}$ for each linkage set $\textbf{F}^i$.
If the average improvement along any of the principal axes of $P^{\mathcal N,i}(\cdot)$ is larger than 1 standard deviation, $c^{mult}_{\textbf{F}^i}$ is increased by $1 / 0.9$.
If no improvements are found, the multiplier is decreased by $0.9$.
The multipliers are prevented to become smaller than 1 for at least a predefined number of generations, the so-called maximum No Improvement Stretch $\text{NIS}^{\text{MAX}}$. For more details, see, e.g.,~\cite{bouter_achieving_2021}.



\subsection{Conditional Linkage Modeling}
\vspace{-1mm}
An approach to detecting pairwise dependencies between variables, including a notion of dependency strength, that has proven to be successful, is fitness-based dependency tests~\cite{munetomo_linkage_1999}.
The Dependency Strength Matrix (DSM)~\cite{hsu_optimization_2015} is a square matrix $\mathbf{D}_{ij} = d(i,j)$, where $i,j \in \{0,1,\ldots,\ell-1\}$ and $d(i,j) \in [0,1]$ indicates dependency strength with 0 indicating no dependence.
Construction of a complete DSM requires $\frac{1}{2}(\ell^2 - \ell)$ fitness dependency tests, imposing a quadratic number of function evaluations as overhead.
However, when partial evaluations can be performed, the overhead of these tests may be reduced to a linear order, depending on the size of the sub-functions.
The DSM can be built incrementally during optimization, which often leads to performing less dependency tests~\cite{andreadis_fitness-based_2024}.

A VIG can be constructed based on the DSM by adding an edge $(u,v)$ whenever $d(u,v) > d_{\mbox{\scriptsize\emph{min}}}$ holds.
In the recent-most work on RV-GOMEA, $d_{\mbox{\scriptsize\emph{min}}}=10^{-6}$ was used and found to lead to good results on a variety of problems~\cite{andreadis_fitness-based_2024}.

In each generation, after an update of the DSM, the VIG must be updated. The VIG can then be used to construct the linkage model that is used in RV-GOMEA. In the recent-most version of RV-GOMEA, a technique called clique seeding is used~\cite{andreadis_fitness-based_2024}.
This technique performs breadth-first searches of cliques in the VIG, starting from each node in the VIG.
For each search, the found maximal clique is stored in a set of candidate cliques, any duplicates are discarded.
Each clique is conditioned on any variables that are outside the clique but have an edge in the VIG to a variable inside the clique.
For each of the candidate cliques a conditional linkage set $\textbf{F}^{i}$ is created, which now describes a \emph{conditional} multivariate distribution. I.e., a conditional linkage set consists of \emph{two} sets of variable indices $\mathcal{I}_{i}^{0}$ and $\mathcal{I}_{i}^{1}$, describing the distribution $P^{\mathcal N}(\vec{X}_{\mathcal{I}_{i}^{0}}|\vec{X}_{\mathcal{I}_{i}^{1}})$.

While conditional linkage sets enable explicitly modeling overlapping dependencies, linkages may still be broken when performing GOM, because acceptance selection is performed immediately after partial variation.
To repair potentially broken linkages, an additional special GOM step is performed each generation in which all variables are sampled without intermediate GOM acceptance, i.e., following the conditionally factorized distribution defined by all conditional linkage sets together. For details, see~\cite{andreadis_fitness-based_2024}.

We store a covariance matrix for each linkage set. If a linkage set is a conditional one, a covariance matrix is stored for all involved variables (i.e., both those in $\mathcal{I}_{i}^{0}$ and in $\mathcal{I}_{i}^{1}$) as the conditional distribution is defined using all covariance values.

\vspace*{-1mm}

\section{iRV-GOMEA}\label{SectioniRVGOMEA}
\vspace*{-1mm}
\subsection{Incremental Distribution Estimation}
We consider two scenarios to incorporate incremental learning into RV-GOMEA: a static scenario where the linkage model is obtained using clique seeding based on the true VIG, and an online scenario where the VIG is obtained through incremental DSM updates using fitness-based dependency tests. Incremental learning is considered for each linkage set in the current generation, in turn, i.e., $\textbf{F}^i_g$.

\begin{figure}[!t]
    \centering
    \includegraphics[width=78mm]{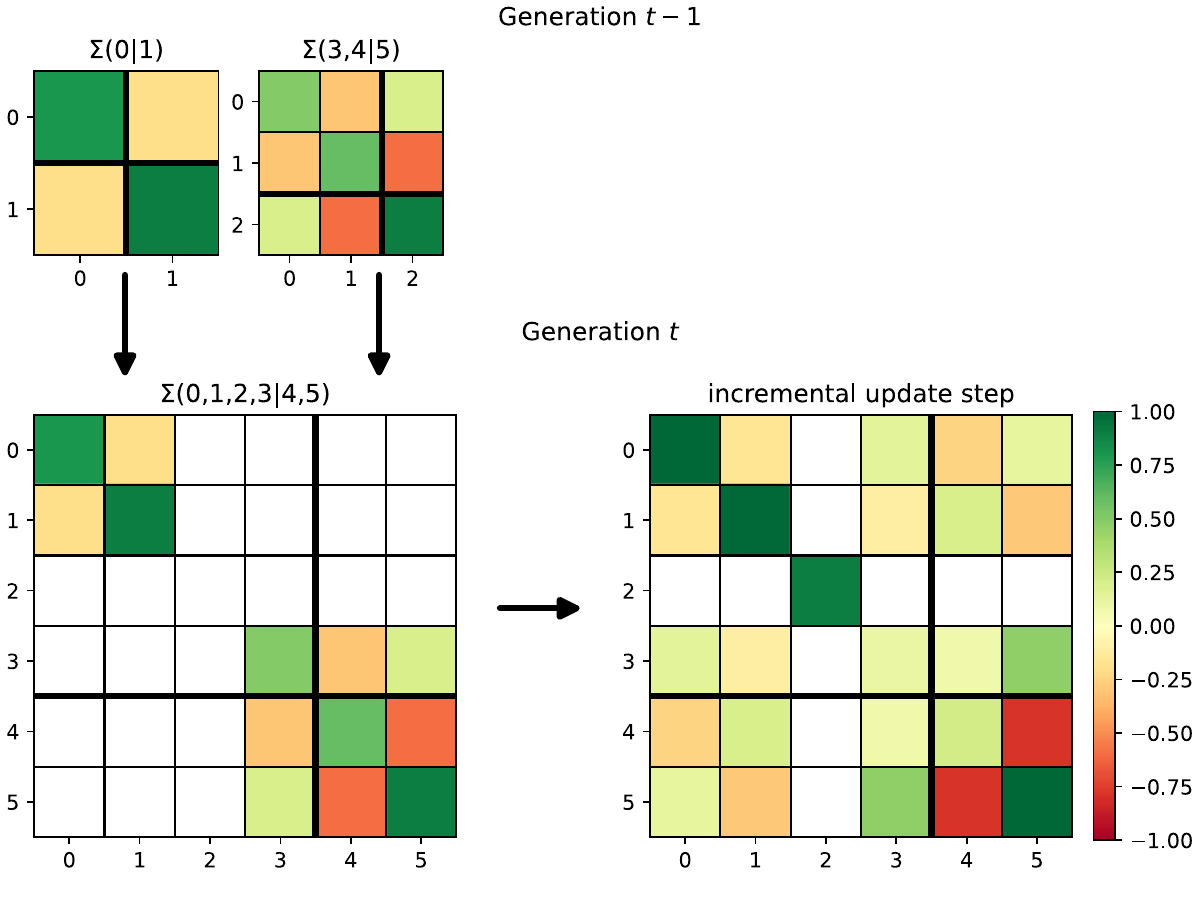}
    \caption{Illustration of copying covariance information between generations and performing a subsequent incremental update step. Colors visualize correlation coefficients.}
    \label{fig:covariance_update}
    \vspace*{-7mm}
\end{figure}

\subsubsection{Covariance Matrices}
In the static scenario, the covariance matrices $\boldsymbol{\Sigma}^{\textbf{F}^i}$ of distributions $P^{\mathcal N,i}(\cdot)$ are initialized with maximum-likelihood estimates of variances from the selection in the first generation on the diagonal and covariance values of 0 off-diagonal.
In subsequent generations, Eq.~\ref{eq:memory} is applied to each $\boldsymbol{\Sigma}^{\textbf{F}^i}$.

In the online scenario, the linkage model changes in each generation. Consequently, the size and involved variables of the covariance matrices $\boldsymbol{\Sigma}^{\textbf{F}^i}$ associated with the linkage sets $\textbf{F}^i$ also change in each generation.
To incrementally update covariances, a method is needed to transfer as much covariance information as possible.

When there exists a linkage set in the previous generation that considers the same variables as used in $\textbf{F}^i_g$, i.e., $\bigcup\textbf{F}^j_{g-1} = \bigcup\textbf{F}^i_g$, then the update step is the same as in the static scenario. I.e., Eq.~\ref{eq:memory} can be used where $\boldsymbol{\Sigma}^{\textbf{F}^i_g}$ is $\vec{\theta}(t)$, $\boldsymbol{\Sigma}^{\textbf{F}^j_{g-1}}$ is used for $\vec{\theta}(t-1)$, and $\hat{\vec{\theta}}(t)$ are the maximum-likelihood estimates from the selection set $\mathcal{S}_g$ in generation $g$ for all (co)variances pertaining to the variables in $\textbf{F}^i_g$.

Otherwise, we proceed as follows:
\begin{enumerate}[leftmargin=8mm]
    \item Initialize a matrix $\boldsymbol{\Sigma}$ of the same dimensions as $\boldsymbol{\Sigma}^{\textbf{F}^i_g}$ to a matrix of zeros and let $\textbf{V}=\bigcup\textbf{F}^i_g$.
    \item Find all possible linkage sets in the previous generation from which information can be reused for the distribution parameters pertaining to $\textbf{F}^{i}_{g}$, i.e., $\mathbb{F} = \left\{\,\textbf{F}^j_{g-1} \in \textbf{F}_{g-1} \ \middle|\ \bigcup\textbf{F}^j_{g-1} \subset \textbf{V}\,\right\}$.
    \item Find the largest set in $\mathbb{F}$ (with ties being broken randomly), i.e.,: $\textbf{F}^{max} = \arg\max_{\textbf{F}^{j}_{g-1} \in \mathbb{F}}\left\{\ \left|\bigcup\textbf{F}^{j}_{g-1}\right|\ \right\}.$
    \item Copy variances and covariances from $\boldsymbol{\Sigma}^{\textbf{F}^{max}}$ to $\boldsymbol{\Sigma}$.
    \item Update $\textbf{V}$ by setting it to $\textbf{V} - \bigcup\textbf{F}^{max}$ and $\mathbb{F}$ to $\mathbb{F}$ - $\textbf{F}^{max}$.
    \item If either either $|\mathbb{F}|$ or $|\textbf{V}|$ is zero, then exit, otherwise, repeat the process, starting at finding $\mathbb{F}$ (step 2).
\end{enumerate}
After this, $\boldsymbol{\Sigma}^{\textbf{F}^i_g}$ can be computed with Eq.~\ref{eq:memory} by using $\boldsymbol{\Sigma}$ for $\vec{\theta}(t-1)$ and the variances and covariances estimated using maximum-likelihood estimates from the selection set $\mathcal{S}_g$ in generation $g$ for $\hat{\vec{\theta}}(t)$. However, for any indices $v$ that remain in $\textbf{V}$, we estimate only the maximum-likelihood variances. Any covariance associated with remaining variables in $\textbf{V}$ are kept at zero. This is similar to the initialization of the covariance matrices in static linkage models.
See Figure \ref{fig:covariance_update} for a visual representation of the described approach.

\subsubsection{Anticipated Mean Shift}
In RV-GOMEA, the anticipated mean shift (AMS), denoted $\boldsymbol{\mu}_{AMS}$ is estimated once for the full set of variables.
For each linkage set $\textbf{F}^i$, the $\boldsymbol{\mu}_{AMS}^{\textbf{F}^i}$ that is used, is equal to the subset of $\boldsymbol{\mu}_{AMS}$ pertaining to the variables in $\textbf{F}^i$. Of note, in case of conditional linkage sets, this pertains only to the conditioned variables, not the ones they are conditioned on.
As was the case for the incremental AMaLGaM~\cite{bosman_empirical_2009}, we aim to apply incremental learning also to the AMS.
Ultimately, the applied AMS depends on the size of the linkage set. However, Equation \ref{eq:learning_rate_function} is dependent on the number of variables.
By incrementally estimating the AMS only for the full set of variables at once, the learning rate may not properly reflect the fact that the AMS is in fact applied in GOM steps that pertain to subsets of variables.
We therefore change this such that in iRV-GOMEA the AMS is also learned independently for each linkage set $\textbf{F}^i$ in the form of $\boldsymbol{\mu}_{AMS}^{\textbf{F}^i}(t)$.



\vspace{-1mm}
\subsection{Distribution Multipliers}
Preliminary experiments with iRV-GOMEA indicated that the direct implementation of incremental learning in RV-GOMEA led to unexpected poor results for small population sizes.
This is counterintuitive because the use of incremental learning has the potential to reduce the minimally required population size.
The number of required generations will then typically go up.
While this balance ultimately is typically in favor of smaller population sizes, resulting in fewer function evaluations required, the number of times the distribution parameters are updated, also goes up.
In RV-GOMEA, different from CMA-ES or AMaLGaM, updates are performed for each linkage set separately and there is a distribution multiplier per linkage set.
Without linkage sets, or for very large linkage sets, multiple variables are changed at the same time and while some changes separately would lead to worse fitness, the joint change could still lead to an improvement. The smaller the linkage sets, the stronger individual variable changes must be correlated with making improvements.
The use of linkage sets in RV-GOMEA may therefore lead to less improvements per generation to be found and the distribution multipliers converting back to the value of 1 quickly after they were increased due to an improvement.
Moreover, with incremental learning, smaller population sizes are made possible, necessitating however more generations before the covariances become aligned with the (local) landscape of the problem being solved and exacerbating the problem of finding improvements per variation of single FOS elements. The increased number of generations also entails more updates to the distribution multipliers being done in iRV-GOMEA (with a smaller population size) than in RV-GOMEA. 
Altogether, the consequence is that the distribution multipliers are reverted back to 1 too quickly, leading to limited numbers of improvements and premature convergence taking place if the population size becomes too small, which contradicts what we are trying to achieve.
For this reason, we changed the distribution multiplier decrease factor to be asymmetrical to the distribution multiplier increase factor, decreasing the value more slowly than it is increased.
Specifically, we set the decrease rate to half the increase rate, i.e.,: 0.95 (and $1/0.9$ for the increase rate).
\vspace{-1mm}
\subsection{Regressing the Learning Rates}
\label{sec:incremental-learning}
To obtain learning rates for each linkage set $\textbf{F}^i$, and for each of the categories of distribution parameters (covariance matrices and AMS), function $f^\eta_{\vec{\alpha}}$ (Eq.~\ref{eq:learning_rate_function}) can be used
where $\ell$ is replaced by the number of variables in $\textbf{F}^i$, which we will denote by $\kappa$ in the remainder of this paper.
To determine proper values for the $\vec{\alpha}$ parameters in $f^\eta_{\vec{\alpha}}$ for both the covariance matrix learning rate and the AMS learning rate, a set of samples $\{((\eta^{\Sigma},\eta^{AMS}), (|\mathbf{\mathcal S}|, \kappa))\}$, is required to perform regression using Equation \ref{eq:learning_rate_fitting}.
These samples must span a large variety of different selection sizes, linkage set sizes, and benchmark problems to obtain a generically usable $f^\eta_{\vec{\alpha}}$.

To create the samples, RV-GOMEA is used to optimize the learning rates $\eta^{\Sigma}$ and $\eta^{\mathrm{AMS}}$ for each problem in Table \ref{tab:lr_opt_problems}. This is done by considering them as variables in a two-dimensional problem that involves running iRV-GOMEA and minimizing the required number of function evaluations to achieve the Value-To-Reach (VTR).

Learning rates are optimized for non-conditional linkage set sizes $\kappa \in \{5,10,15,...,100\}$ and population sizes $|\mathbf{\mathcal P}| \in \{20,40,60,...,400\}$, with the selection percentile the same as in AMaLGaM and RV-GOMEA, $\tau=0.35$ ($|\mathbf{\mathcal S}| = \lfloor\tau|\mathbf{\mathcal P}|\rfloor$).
Solutions in iRV-GOMEA are initialized in $[-10, 5]$, a range located asymmetrically around the optimum (at 0).
For each optimization problem and for each value of $\kappa$ and $|\mathbf{\mathcal P}|$, iRV-GOMEA is given a budget of twice the number of evaluations that are required to reach the VTR using the non-incremental RV-GOMEA with a full covariance matrix, for which the guideline population size of $3\kappa^{1.5} + 17$ is used \cite{bosman_empirical_2009}.
If iRV-GOMEA exceeds that budget, the sample $((\eta^{\Sigma},\eta^{AMS}), (|\mathbf{\mathcal S}|, \kappa))$ for that problem is discarded.

(i)RV-GOMEA reverts to univariate sampling of the distribution when the (incrementally) estimated covariance matrix is singular.
This can happen when the selection size $|\mathbf{\mathcal S}|$ is simply too small for a linkage set size $\kappa$.
Since the goal is to determine the learning rates for the correct sampling of full covariance matrices, one of the included optimization problems is the ill-conditioned rotated ellipsoid problem.
Reverting to univariate sampling will cause iRV-GOMEA to no longer be able to solve this problem. Moreover, when iRV-GOMEA fails to find the optimum for the rotated ellipsoid problem of a certain linkage set size $\kappa$ and with a population size $|\mathbf{\mathcal P}|$, all samples for other problems with linkage set size $\kappa$ and population size $|\mathbf{\mathcal P}|$ are removed from the set of samples.

The resulting set of samples for both $\eta^{\Sigma}$ and $\eta^{\mathrm{AMS}}$ are used as input for Equation \ref{eq:learning_rate_fitting}.
The problem defined by this equation is then optimized using RV-GOMEA to instantiate $f^\eta_{\vec{\alpha}}$ for both the AMS and covariance matrix separately.

\begin{table}[h]
\vspace*{-4mm}
\centering
\caption{Optimization problems used in the experiments.}
\vspace*{-1.5mm}
\label{tab:lr_opt_problems}
\setlength\tabcolsep{3.5pt}
\renewcommand{\arraystretch}{1.3}
\scalebox{1.0}
{
\scriptsize
\begin{tabular}{|l|l|l|}
\hline
Problem          & Function Definition                                                         & VTR        \\ \hline
Sphere           & $\sum_{i=0}^{\ell-1} x_i^2$                                                    & $10^{-10}$ \\[0.5mm]
Rotated Ellipsoid& $\sum_{i=0}^{\ell-1} 10^{6\frac{i}{\ell-1}} (x^r_i)^2$, where $\vec{x^r} = \vec{R}\vec{x}$ for rotation matrix $\vec{R}$ & $10^{-10}$ \\[1mm]
Cigar            & $x_0^2 + \sum_{i=1}^{\ell-1} 10^6 x_i^2$                                         & $10^{-10}$ \\[1mm]
Tablet           & $10^6 x_0^2 + \sum_{i=1}^{\ell-1} x_i^2$                                         & $10^{-10}$ \\[1mm]
Cigar Tablet     & $x_0^2 + \sum_{i=1}^{\ell-2} 10^4 x_i^2 + 10^8x_{\ell-1}^2$                         & $10^{-10}$ \\[1mm]
Two Axes         & $\sum_{i=0}^{\floor{\ell/2}-1} 10^6 x_i^2 + \sum_{i=\floor{\ell/2}-1}^{\ell-1}x_i^2$ & $10^{-10}$ \\[1mm]
Different Powers & $\sum_{i=0}^{\ell-1} |x_i|^{2+10\frac{i}{\ell-1}}$                               & $10^{-10}$ \\[1mm]
Rosenbrock       & $\sum_{i=0}^{\ell-2} \left(100 (x_i^2 -x_{i+1})^2 + (x_i -1)^2\right)$         & $10^{-10}$ \\[1mm]
Parabolic Ridge  & $-x_1 + 100\sum_{i=0}^{\ell-1} x_i^2$                                          & $-10^{10}$ \\[1mm]
Sharp Ridge      & $-x_1 + 100 \sqrt{\sum_{i=0}^{\ell-1} x_i^2}$                                  & $-10^{10}$ \\[1mm] \hline
\end{tabular}
}
\vspace*{-2mm}
\end{table}

\begin{figure}[h!]
    \vspace*{-2.5mm}
    \begin{minipage}{\columnwidth}
      \centering
        \includegraphics[width=80mm]{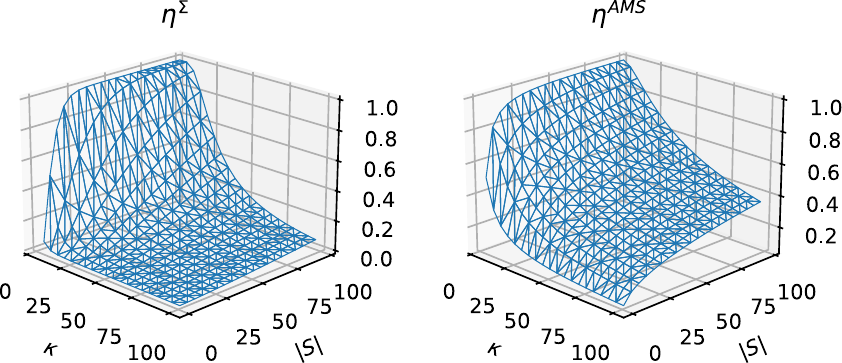}
      \vspace*{0mm}
      \caption{Regressed $f^\eta_{\vec{\alpha}}$ functions over all problems.}
      \label{fig:learning_rates_bbo}
    \end{minipage}%
    \hfill
    \begin{minipage}{0.32\textwidth}
        \centering
        \captionof{table}{Regressed $\vec{\alpha}$ values over all problems.}
        \label{tab:alphas_bbo}
        \vspace{-1.5mm}
        \setlength\tabcolsep{3.5pt}
        \renewcommand{\arraystretch}{1.2}
        \begin{tabular}[t]{c|c|c|c}
                 & $\alpha_0$ & $\alpha_1$ & $\alpha_2$ \\ \hline
        $\eta^{\Sigma}$ & -1.01      & 1.32       & 1.94       \\ \hline
        $\eta^{AMS}$      & -2.95      & 0.47       & 0.87      
        \end{tabular}
        \vspace*{-5mm}
    \end{minipage}
\end{figure}

The regressed $\vec{\alpha}$ parameters, after performing the above for all problems in Table~\ref{tab:lr_opt_problems} and merging all samples into one dataset, can be found in Table \ref{tab:alphas_bbo}.
A visualization of the learning rates $\eta^\Sigma$ and $\eta^\mathrm{AMS}$ can be found in Figure \ref{fig:learning_rates_bbo}. Using the regressed learning rate formulas in the incremental distribution estimation procedure as outlined above within RV-GOMEA, the final iRV-GOMEA is obtained. Next, we turn to experimenting with the now fully defined iRV-GOMEA.

\subsection{Determining Population Sizing Guidelines for a Single Linkage Set in iRV-GOMEA}
One of the main motivations for incremental distribution estimation is the potential for requiring smaller population sizes to solve problems.
We determine the required population size for each problem in Table \ref{tab:lr_opt_problems} for linkage set sizes up to $\kappa = 200$. Note that in iRV-GOMEA now a single linkage set is used, and as such, a full covariance matrix is estimated over all involved problem variables, regardless of whether the problem is actually decomposable.
Beginning at the default population size of RV-GOMEA, the population size is halved as long as the required number of evaluations (averaged over 100 runs) to achieve the VTR, reduces. 
Once the required number of evaluations stops decreasing, binary search is performed between the last two tried population sizes, until the best performing population size is found.
The required number of evaluations is divided by the success probability to account for failed runs with very small population sizes. 

\begin{figure*}[ht]
    \vspace*{-5mm}
    \centering
    \begin{subfigure}[t]{0.41\textwidth}
        \centering
        \includegraphics[width=\textwidth]{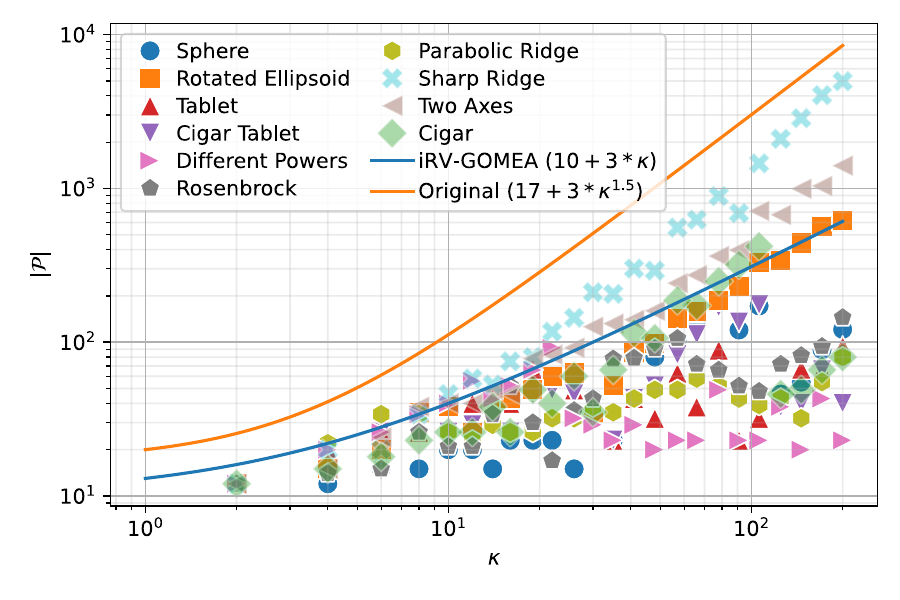}
        \label{fig:popsize}
    \end{subfigure}
    \begin{subfigure}[t]{0.41\textwidth}
        \centering
        \includegraphics[width=\textwidth]{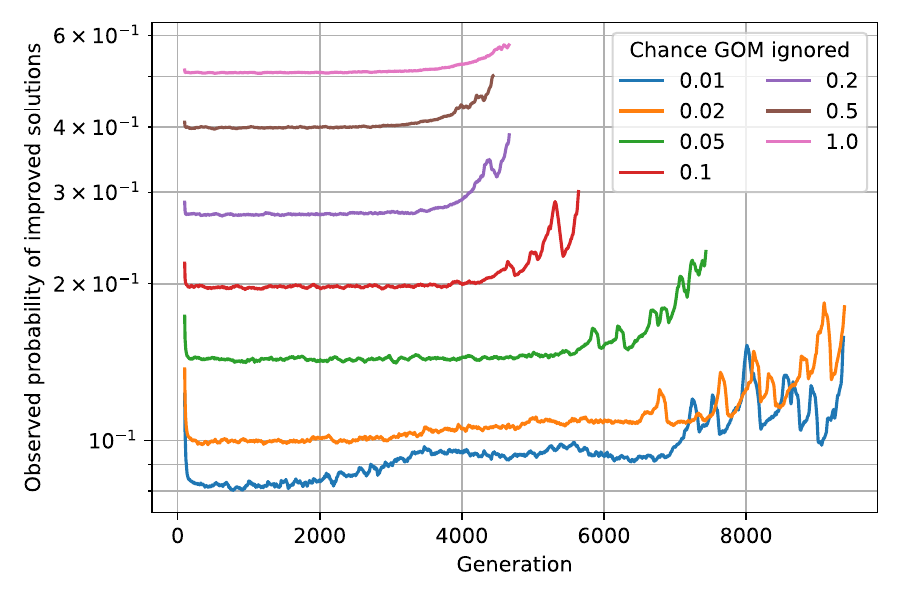}
        \label{fig:gom_selection}
    \end{subfigure}
    \vspace*{-7mm}
    \caption{Population size guideline results. Left: Best performing population size $|\mathcal{P}|$ per problem and linkage set size $\kappa$ (symbols) and guidelines (solid lines). Right: Observed probability of improvement when ignoring GOM on Sharp Ridge with different probabilities.}
    \label{fig:gom_popsize_and_selection}
    \vspace*{-5mm}
\end{figure*}

The best performing population size for each problem and linkage set size $\kappa$ can be seen in Figure \ref{fig:gom_popsize_and_selection}. We want to use these results to derive a population size guideline. 
Sharp Ridge results were however ignored when determining the guideline, as deviating behavior was observed.
Specifically, the criterion to only accept improvements in GOM seems to prevent finding improvements over multiple generations.
The fraction of the population that improves if the acceptance criterion in GOM is ignored with increasing probability (to the point where all samples are always accepted, which corresponds to classical EDA configurations as used in AMaLGaM and CMA-ES), can be seen in Figure \ref{fig:gom_popsize_and_selection} on Sharp Ridge with $\kappa = 100$.
Clearly, when the acceptance selection in GOM is ignored more frequently, the optimum is found in fewer generations. While this can be considered a shortcoming, it is important to note that this holds only when using the full covariance matrix on a (sub)function of $\kappa$ variables with $\kappa$ fairly large. Arguably, oftentimes, problems have subfunctions pertaining to smaller sets of fully dependent variables that should be considered jointly. Moreover, the observed limitation holds for a set of variables in the Sharp Ridge function, the variables of which are actually not interdependent. As such, these variables in practice should be modeled as independent and be in separate linkage sets of size $\kappa=1$. For similar reasons, we also disregarded the results on Two Axes, for which similar behavior was observed. Finally, for various problems, e.g., Cigar, a drop in population size can clearly be identified in Figure~\ref{fig:gom_popsize_and_selection} as $\kappa$ goes up. This is because there is a large basin of approximately equally-well performing population sizes (for an example, see supplementary material A.2), but we have automated finding the best performing population size. Ultimately, we therefore opted to fit a function to the upper envelope of the problem that leads to the largest population size without deviating behavior, which was Rotated Ellipsoid. Based on that function, we determined guideline for the required population size for linkage set sizes $\kappa$ in iRV-GOMEA to be:
\begin{equation}
    10 + 3 \kappa.
    \label{eq:popsize_irv}
\end{equation}

\vspace*{-5mm}
\section{Experiments}\label{SectionExperiments}
In this section, we describe the various experiments that we performed with the finally obtained iRV-GOMEA. First, we consider the use of a single, static, non-conditional linkage set in iRV-GOMEA and compare performance with the non-incremental RV-GOMEA and RV-GOMEA$_{C}$, the CMA-ES variant of RV-GOMEA. This can be considered a classical black-box evaluation setting.
We subsequently evaluate the performance of iRV-GOMEA when the number of static non-conditional linkage sets is scaled up and partial evaluations can be performed, i.e., a gray-box setting. We compare again with RV-GOMEA and RV-GOMEA$_{C}$, but also with VkD-CMA-ES~\cite{akimoto_projection-based_2016,akimoto_online_2016}, a variant of CMA-ES that learns to decompose the Gaussian model, however in a manner that cannot leverage partial evaluations. Lastly, we consider again the gray-box setting, and evaluate iRV-GOMEA with \emph{conditional} linkage sets, defined both statically and via online fitness-based linkage learning, on a variety of problems and we compare its performance to RV-GOMEA and VkD-CMA-ES. Overview of algorithm hyperparameters can be found in supplementary material A.1.
Source code of iRV-GOMEA is provided on Github\footnote{\href{https://github.com/renzoscholman/irv-gomea}{https://github.com/renzoscholman/irv-gomea}}.

\subsection{Single Non-Conditional Linkage Sets}
We conducted experiments for a single linkage set of size $\kappa = \ell$ up to 200 variables for each problem in Table~\ref{tab:lr_opt_problems}.
Such sizes likely cover most real-world scenarios where the maximum number of fully dependent variables is expected to be smaller~\cite{andreadis_morea_2023,luong_application_2018,chen_tunneling_2018,cai_wind_2023}.

We consider iRV-GOMEA as defined in this paper, the non-incremental RV-GOMEA and RV-GOMEA$_{C}$, and VkD-CMA-ES.
The population sizes are set to their guideline values:
\begin{itemize}
\item For iRV-GOMEA we followed Equation~\ref{eq:popsize_irv} with $\ell=\kappa$.
\item For RV-GOMEA we used $3\ell^{1.5} + 17$ ~\cite{bosman_empirical_2009}.
\item For RV-GOMEA$_C$ and VkD-CMA-ES we used $4 + \lfloor3 * \ln(\ell)\rfloor$, following guidelines for the CMA-ES         sampling model ~\cite{hansen_completely_2001,akimoto_online_2016}.
\end{itemize}
Solutions are uniformly randomly initialized in $\left[-115, -100\right]$ and the required number of evaluations to reach the VTR is averaged over 100 runs.

\subsection{Scaling Multiple Linkage Sets}
All algorithms are tested using static non-conditional linkage models on the Sphere and Rosenbrock problem as well as on the SoREB problem, which is derived from the Rotated Ellipsoid Blocks (REB), that is defined as follows:
$$f^{REB}(\textbf{x},c,\theta,\kappa,s) = \sum_{i=0}^{\left\lceil{\frac{|\textbf{x}| - \kappa}{s}}\right\rceil - 1} f^{ellipsoid}(R^\theta(x_{\lceil{is:is+\kappa-1}\rceil}),c)$$
\vspace*{-3mm}
$$f^{ellipsoid}(\textbf{x},c) = \sum_{i=0}^{|\textbf{x}| - 1} 10^{c\frac{i}{|\textbf{x}|-1}} x_i^2$$
Here, $c$ is the condition number of a subfunction, $\theta$ the rotation angle, $\kappa$ the block size, and $s \geq 1$ the block stride. The SoREB problem is now defined as $f^{SoREB}(\textbf{x},\kappa) = f^{REB}(\textbf{x}, 6, 45, \kappa, \kappa)$, i.e., it is a sum of $\kappa$-dimensional consecutively encoded rotated ellipsoids.

Again, solutions are initialized in $\left[-115, -100\right]$.
For SoREB, block sizes $\kappa \in \{5,10,15,20,25,30\}$ are used for $\ell$ up to $2000$.
For RV-GOMEA the population size guideline of $3\kappa^{1.5} + 17$ is used, i.e., based on the block size.
Similarly for iRV-GOMEA and RV-GOMEA$_C$, the block size $\kappa$ is used instead of $\ell$ for the population guideline.
For VkD-CMA-ES, the population size is kept at $4 + \lfloor3 * \ln(\ell)\rfloor$, conforming with literature~\cite{akimoto_projection-based_2016,akimoto_online_2016}.
For Rosenbrock, problem sizes up to $\ell = 10^4$, and for Sphere up to $\ell = 10^5$, are tested.
A univariate linkage model is used for Sphere.
For Rosenbrock, a linkage model with non-conditional, but overlapping blocks of 2 is used defined according to the additive composition of the problem definition.
We investigate the average number of required evaluations to reach the VTR, which we set to $10^{-10}$, over 100 runs.

\subsection{Gray-Box Optimization}
For online Fitness-Based (FB) linkage learning, we employ the method from~\cite{andreadis_fitness-based_2024}.
We further consider static conditional linkage models based on the true VIG.
For both cases, we use the recently introduced Clique Seeding (CS) method~\cite{andreadis_fitness-based_2024} to maximize the size of the conditional linkage sets.
These two cases are denoted as: "FB-CS-LM" and "Static-CS-LM".

We consider various GBO problems, both separable and non-separable. 
Besides Sphere and Rosenbrock, as seen in Table \ref{tab:lr_opt_problems}, the
remaining problems are defined as \cite{andreadis_fitness-based_2024}:
\begin{equation*}
\begin{aligned}
\ \\[-5mm]
f^{REB2Weak}(\textbf{x})           & = f^{REB}(\textbf{x}, 1, 5, 2, 1) \\[-1.5mm]
f^{REB2Strong}(\textbf{x})         & = f^{REB}(\textbf{x}, 6, 5, 2, 1) \\[-1.5mm]
f^{REB2Alternating}(\textbf{x})    & = f^{REB}(\textbf{x}, c_i, \theta_i, 2, 1) \\[-1.5mm]
f^{REB5NoOverlap}(\textbf{x})      & = f^{REB}(\textbf{x}, 6, 45, 5, 0) \\[-1.5mm]
f^{REB5SmallOverlap}(\textbf{x})   & = f^{REB}(\textbf{x}, 6, 45, 5, 1) \\[-1.5mm]
f^{REB5LargeOverlap}(\textbf{x})   & = f^{REB}(\textbf{x}, 6, 45, 5, 4) \\[-1.5mm]
f^{REB5Alternating}(\textbf{x})    & = f^{REB}(\textbf{x}, c_i, \theta_i, 5, 4) \\[-1.5mm]
    \end{aligned}
    \end{equation*}
    \begin{equation*}
    \begin{aligned}
f^{REB5DisjointPairs}(\textbf{x})  & = f^{REB}(\textbf{x}, 6, 45, 5, s_i) \\[-1.5mm]
f^{REB10NoOverlap}(\textbf{x})     & = f^{REB}(\textbf{x}, 6, 45, 10, 0) \\[-1.5mm]
f^{REB10SmallOverlap}(\textbf{x})  & = f^{REB}(\textbf{x}, 6, 45, 10, 1) \\[-1.5mm]
f^{REB10LargeOverlap}(\textbf{x})  & = f^{REB}(\textbf{x}, 6, 45, 10, 4) \\[-1.5mm]
f^{REB10Alternating}(\textbf{x})   & = f^{REB}(\textbf{x}, c_i, \theta_i, 10, 4) \\
\end{aligned}
\end{equation*}

For the alternating problems, $c_i = 1$ and $\theta_i = 5$ if $i$ (block index) is even, else  $c_i = 6$ and $\theta_i = 45$.
For the disjoint pair problems, $s_i = 4$ if $i$ is even, else $s_i = 5$.

Lastly, we use the OSoREB problem with two degrees of overlapping REB functions and the REBGrid problem.
The REBGrid problem arranges the $\ell$ variables as vertices $v \in V$ in a square grid of dimensions $\sqrt{\ell} * \sqrt{\ell}$, and horizontally and vertically connects neighboring vertices $a \in N(b)$ through an edge. 
These two functions are defined as:
$$f^{OSoREB}(\textbf{x}) = f^{REB}(\textbf{x}, 6, 45, 5, 4) + f^{REB}(x_{[f:|\textbf{x}|-1]}, 6, 45, 2, 5)$$
$$f^{REBGrid}(\textbf{x}) = \sum_{v \in V} f^{ellipsoid}(R^{45}(x_{v \cup N(v)}), 6)$$

For each of the problems above, iRV-GOMEA, RV-GOMEA, and VkD-CMA-ES are tested for $\ell \in \left[10,20,40,80,160,320\right]$.
Some of the REB problems are not defined for these dimensionalities.
In those cases, a proper value for $\ell$ was used that was as similar as possible.
All algorithms have an evaluation budget of $10^8$ to reach the VTR of $10^{-10}$.

Population sizes are optimized for each algorithm, problem instance, and linkage model using bisection search up to $\ell = 80$.
The bisection search is performed similar to how the general population guideline was determined.
Due to large computation time requirements for bisection on larger dimensionalities ($160, 320$), the population size results are extrapolated for those.
This allows testing whether incremental learning can still lead to improvements even when fully tuning the algorithms to a problem.

For iRV-GOMEA and RV-GOMEA both conditional linkage learning models are compared.
In the comparison of iRV-GOMEA to VkD-CMA-ES, only the FB linkage learning is used so that no information is provided to the algorithm a priori and both algorithms learn the problem decomposition during optimization.

\section{Results}\label{SectionResults}
\begin{figure*}[!ht]
    \vspace*{-5mm}
    \centering
        \includegraphics[width=0.86\textwidth]{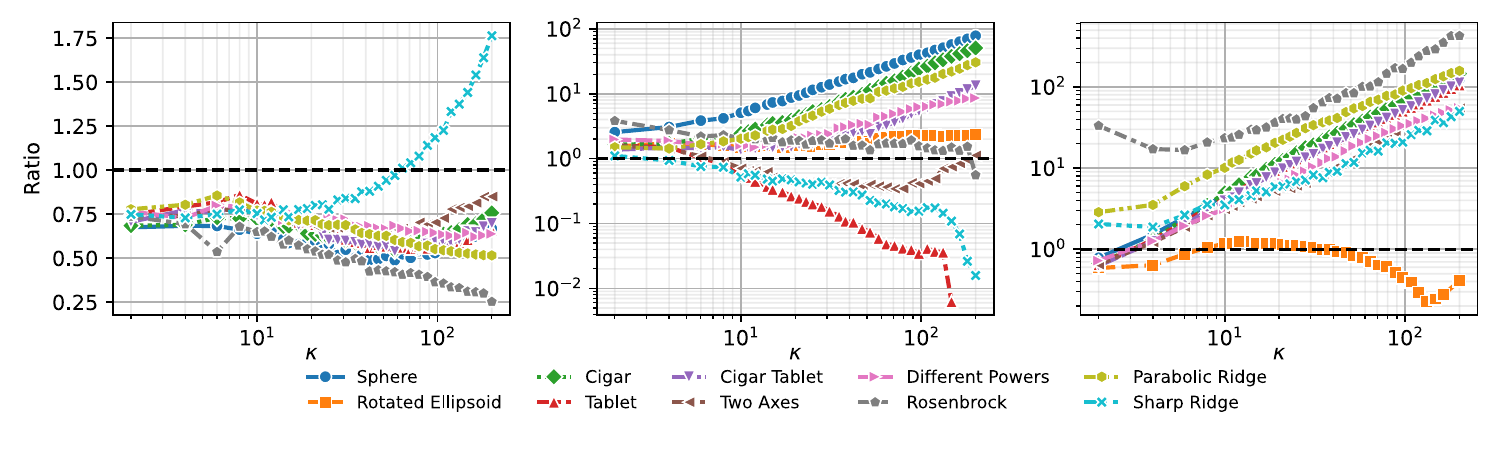}
    \vspace*{-5mm}
    \caption{Varying linkage set size results for iRV-GOMEA vs (respectively) RV-GOMEA, RV-GOMEA$_C$, and VkD-CMA-ES.}
    \label{fig:results-bbo}
    \vspace*{-5mm}
\end{figure*}

\subsection{Single Non-Conditional Linkage Sets}
The results for varying the maximum linkage set size $\kappa$ can be seen in Figure \ref{fig:results-bbo}. We report the ratio of the required evaluations between iRV-GOMEA and the other two algorithms ($< 1$ indicates iRV-GOMEA was better). In the supplementary material A.2 we show the required evaluations to reach the VTR.

For most problems, the required number of evaluations is decreased notably when comparing iRV-GOMEA to RV-GOMEA.
However, for larger $\kappa$ the ratio deteriorates, especially for Sharp Ridge.
These results are in contrast to those found for iAMaLGAM versus AMaLGaM \cite{bosman_empirical_2009}, in which GOM is not used.
For larger linkage sets, the larger population size in the guideline of RV-GOMEA works better with the GOM operator, as more opportunities arise in each generation to find improvements.
In practice, this will most likely not have many implications since linkage sets larger than several tens of parameters are not expected to typically be present.

The CMA-ES sampling model in RV-GOMEA$_C$ leads to better performance than iRV-GOMEA on various problems. This is expected, as in this setting of a single linkage set we are essentially comparing iAMaLGaM and CMA-ES in a black-box setting and CMA-ES is the state-of-the-art in this case.
For Sharp Ridge, GOM was already shown to be a limiting factor that prevents improvements. This is reflected again here in the results of all three RV-GOMEA variants.
\vspace{-1mm}

\subsection{Scaling Multiple Linkage Sets}
\begin{figure*}[!ht]
    \centering
        \includegraphics[width=0.77\textwidth]{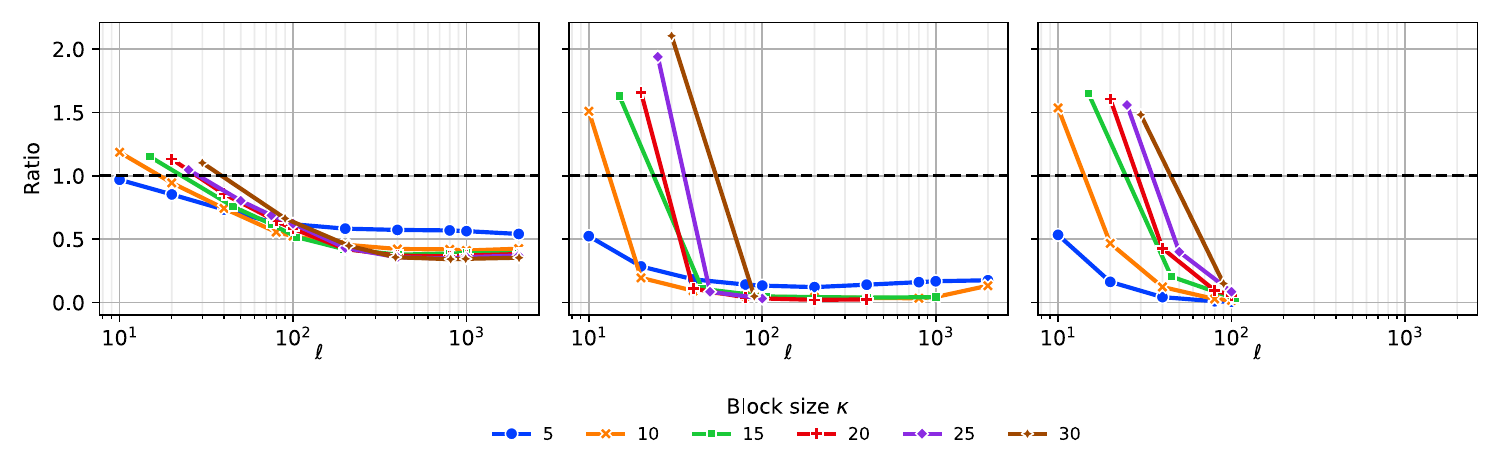}
    \vspace*{-2mm}
    \caption{Gray-box SoREB results iRV-GOMEA vs (respectively) RV-GOMEA, RV-GOMEA$_C$, and VkD-CMA-ES.}
    \label{fig:results-soreb}
\end{figure*}

RV-GOMEA is known to perform really well in GBO settings, even outperforming RV-GOMEA$_C$~\cite{bouter_achieving_2021}. 
Adding incremental learning following the approach in this paper, can lead to even better results, as can be seen in Figure \ref{fig:results-soreb}.
On the SoREB problem, iRV-GOMEA outperforms RV-GOMEA substantially by reaching the VTR two to three times as fast (for problems with more than one block).
Compared to RV-GOMEA$_C$ and VkD-CMA-ES, the improvements are even larger.
As $\ell$ is increased beyond a hundred variables, VkD-CMA-ES starts to give out compared to the scalability of the RV-GOMEA variants.
Even with its decomposition capabilities, it converged prematurely above $\ell = 100$ in our experiments.
In the supplementary material A.3 we show the required evaluations to reach the VTR.
\vspace{-1mm}

\begin{figure*}[!ht]
    \centering
    \vspace*{-3mm}
    \scalebox{0.98}{
    
        \includegraphics[width=0.9\textwidth]{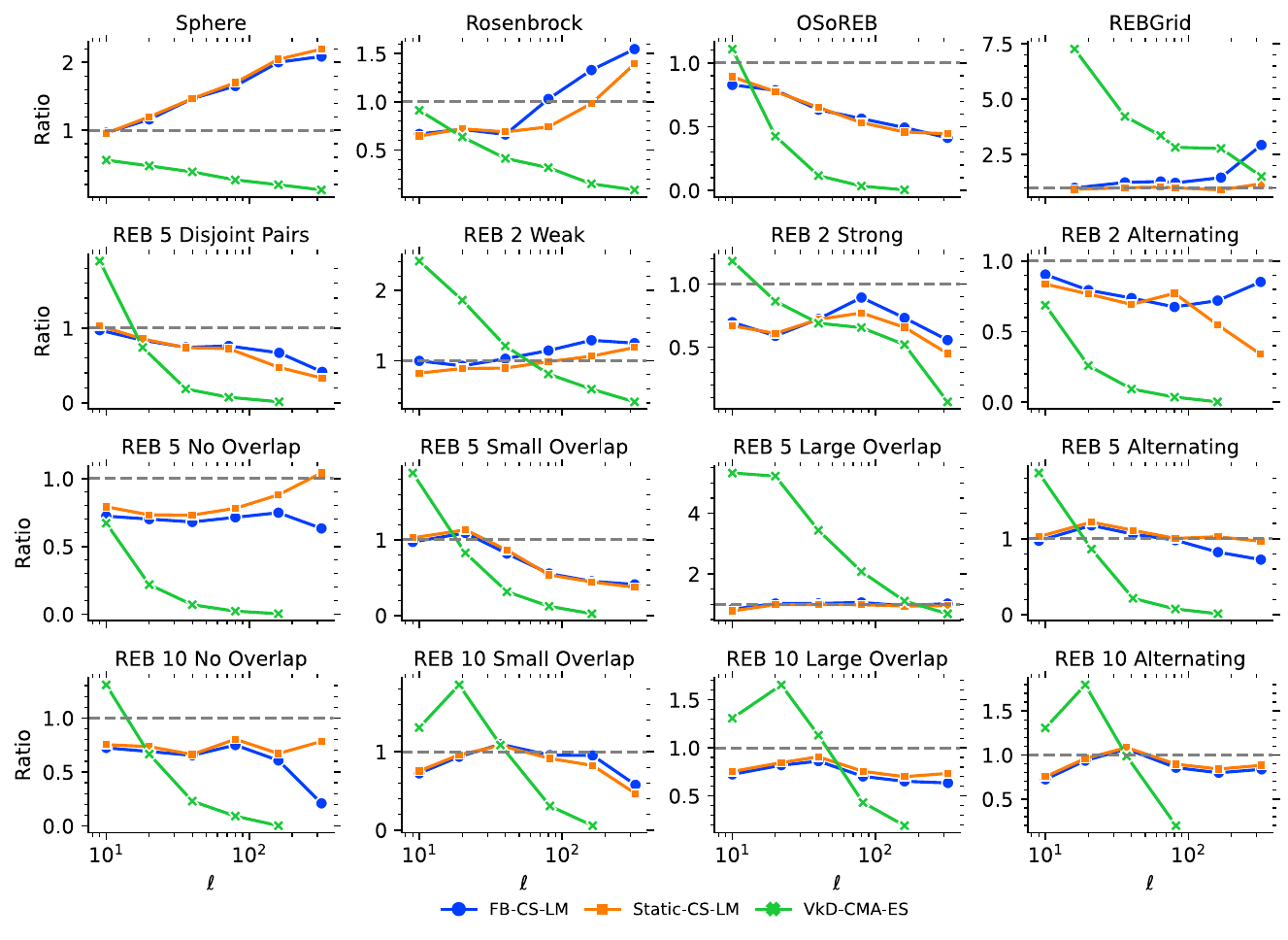}
    }
    \vspace*{-3mm}
    \caption{Ratio of required number of function evaluations to reach the VTR for iRV-GOMEA vs RV-GOMEA and for iRV-GOMEA vs VkD-CMA-ES on all tested GBO benchmark problems. A ratio below 1.0 means that iRV-GOMEA has superior performance.}
    \label{fig:results-conditional}
    \vspace*{-5mm}
\end{figure*}

\subsection{Gray-Box Optimization}
Figure \ref{fig:results-conditional} shows the resulting reduction in the required number of evaluations for iRV-GOMEA versus RV-GOMEA for both linkage models when solving every benchmark problem in a GBO setting and additionally finding the optimally performing population size for each problem dimensionality.
On nearly all problems, the required number of evaluations is reduced.
The optimized population sizes and required evaluations can be found in the supplementary material A.4.

The only case where iRV-GOMEA is clearly worse is the fully separable Sphere problem.
There, the use of the GOM operator with univariate linkage sets allows for very small population sizes to be used also in RV-GOMEA, making incremental learning superfluous.

On average, iRV-GOMEA can be observed to achieve equal performance compared to VkD-CMA-ES when problem dimensionalities are small and there are few subfunctions.
However, as problem dimensionality increases and the relative utility of partial evaluations grows, iRV-GOMEA demonstrates a substantial reduction in the number of required evaluations.
\section{Discussion}\label{SectionDiscussion}
Many real-world problems require the exploitation of problem-specific knowledge to reach good results within a reasonable amount of time. Of particular value can be the ability to exploit problem decomposition together with the ability to perform partial evaluations. While RV-GOMEA is already state-of-the-art and capable of outperforming traditional EAs and EDAs that cannot exploit partial evaluations, the incremental learning of Gaussian distributions had not yet been studied within the GOMEA framework, nor applied to a GBO scenario where partial variation and evaluations are possible.
Therefore, in this paper we have investigated for the first time how incremental learning can be incorporated in RV-GOMEA by empirically discovering a formula for learning rates to be used when incrementally learning covariances and the AMS in RV-GOMEA.

The use of incremental learning as proposed in this paper, improves the performance of RV-GOMEA in many cases. However, 
when linkage set sizes increase, the performance advantage of iRV-GOMEA over RV-GOMEA and RV-GOMEA$_C$ decreases. 
Further research would be required to determine the root cause of this how and that could be tackled. Likely, the main reason is the fact that solutions may not change during GOM (i.e., if changes are rejected) and incremental learning then is poised to learn from the same solutions, deteriorating the covariance learning process.

While the approach taken in this paper to arrive at learning rate formulas for the covariance matrix and the AMS has proven mostly successful, the approach does involve tailoring the learning rate formulas to a set of benchmark problems. For a real-world problem, especially if is a recurring one where instances need to be solved repeatedly, it may work well to use the same setup but then for the specific problem at hand. Additional performance improvements could then well be obtained, which could have important real-world implications. It would be interesting to explore this in future work.

\section{Conclusion}\label{SectionConclusion}
In this paper, we have introduced iRV-GOMEA, a new variant of RV-GOMEA that employs generation-wise incremental learning of the parameters of the sampling model. On several GBO problems with varying dependency structures, iRV-GOMEA was shown to be capable of outperforming RV-GOMEA, RV-GOMEA$_C$, and VkD-CMA-ES. When establishing and using general population-sizing guidelines, the required number of function evaluations could be reduced by a factor of 2 to 3 times when using iRV-GOMEA instead of RV-GOMEA.
Even when tuning population sizes specifically for a problem and dimensionality, the use of incremental learning in RV-GOMEA could still lead to a reduction in the required number of evaluations.
While incremental learning was not found to improve the performance of RV-GOMEA on fully decomposable (i.e., univariate) problems, when larger sets of interdependent variables are present in a problem, the efficiency gains of using incremental learning were clear, both in case static linkage models are provided, and when fitness-based linkage learning is employed.




\begin{acks}
This work was supported by the Dutch Cancer Society [KWF Kankerbestrijding; project number 12183] and by Elekta Brachytherapy, Veenendaal, The Netherlands. 
This work used the Dutch national e-infrastructure with the support of the
SURF Cooperative using grant no. EINF-12015.
\end{acks}

\bibliographystyle{acm}
\bibliography{gecco}
\clearpage
\appendix
\section{Supplementary}

\subsection{Algorithm Hyperparameters}
For the sake of reproducibility, all hyperparameters of the RV-GOMEA variants are denoted in Table \ref{tab:hyperparam}.
For VkD-CMA-ES, the source code was obtained from one of the authors (\href{https://gist.github.com/youheiakimoto/2fb26c0ace43c22b8f19c7796e69e108}{Github}) and hyperparameters were left unchanged.
Only the VTR and maximum number of evaluations (for which settings are provided in the paper) were altered to correctly evaluate and compare it to the RV-GOMEA variants.
\begin{table}[h]
\centering
\scriptsize
\begin{tabular}{llll}
Hyperparameter                     & iRV-GOMEA     & RV-GOMEA            & RV-GOMEA$_C$                      \\ \hline
Population Size Guideline          & $10 + 3 \ell$ & $17 + 3 \ell^{1.5}$ & $4 + \lfloor3 * \ln(\ell)\rfloor$ \\
Number of populations              & 1             & 1                   & 1                                 \\
GOM accept chance                  & 0\%           & 0\%                 & 0\%                               \\
Distribution Multiplier Increase   & $1/0.9$       & $1/0.9$             & N/A                               \\
Distribution Multiplier Decrease   & $0.95$        & $0.9$               & N/A                               \\
Selection percentile $\tau$        & $0.35$        & $0.35$              & $0.5$  (only for AMS)                            \\
Standard Deviation Ratio Threshold & 1.0           & 1.0                 & N/A                               \\
Fitness Variance Tolerance         & 0.0           & 0.0                 & 0.0                               \\
NIS$^{MAX}$                        & 25 + $\ell$   & 25 + $\ell$         & 25 + $\ell$
\end{tabular}
\caption{Used hyperparameters for RV-GOMEA variants.}
\label{tab:hyperparam}
\end{table}

\subsection{Single Non-Conditional Linkage Sets}
\begin{figure*}
    \centering
        \includegraphics[width=\textwidth]{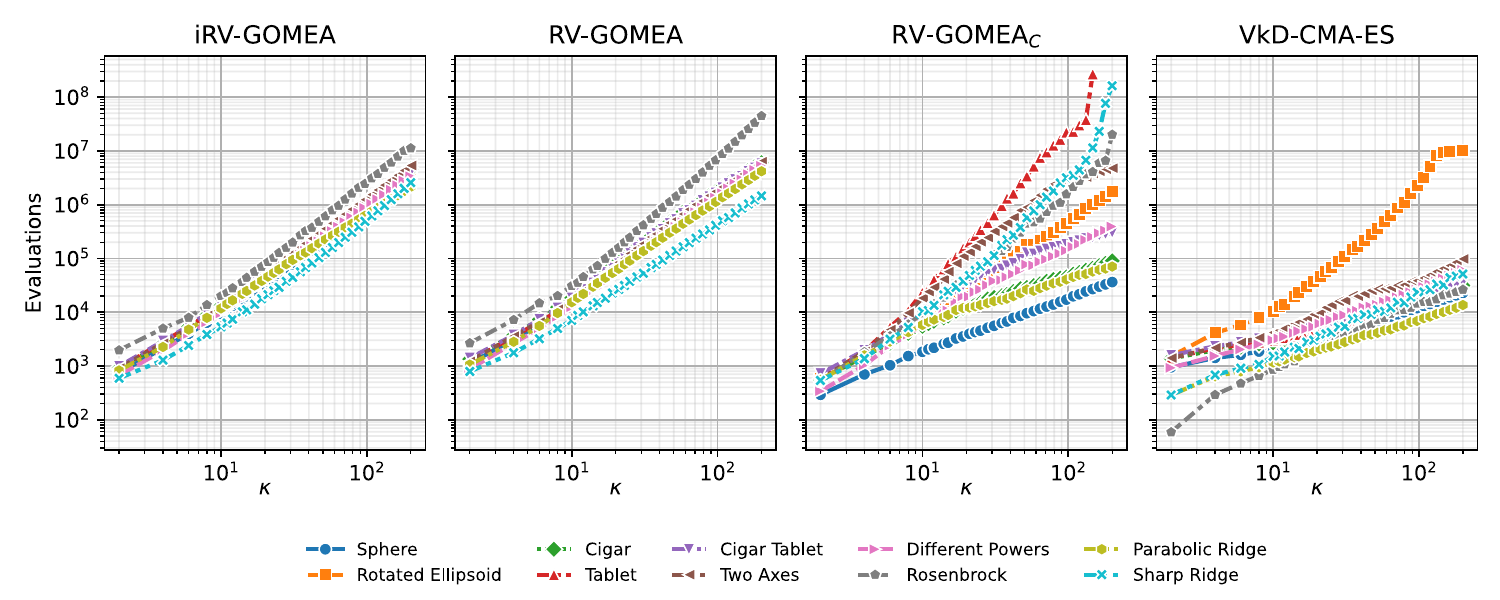}
    \caption{Required number of evaluations to reach VTR for all tested optimization problems with population guidelines based on $\kappa$.}
    \label{fig:req_evals_bbo}
\end{figure*}

The reported ratios of required evaluations for iRV-GOMEA, RV-GOMEA, RV-GOMEA$_C$, and VkD-CMA-ES, were based on the results visualized in Figure \ref{fig:req_evals_bbo}.
These results reiterate that the CMA-ES sampling model is state-of-the-art in black-box optimization.
However, in combination with the GOM operator, the CMA-ES sampling model can also deteriorate in performance, similar to the to the (i)AMaLGaM sampling model employed in (i)RV-GOMEA.
VkD-CMA-ES does seem to perform notably worse for the SoREB problem as compared to all others when the dimensionality increases.

\begin{figure*}[h!]
    \centering
        \includegraphics[width=70mm]{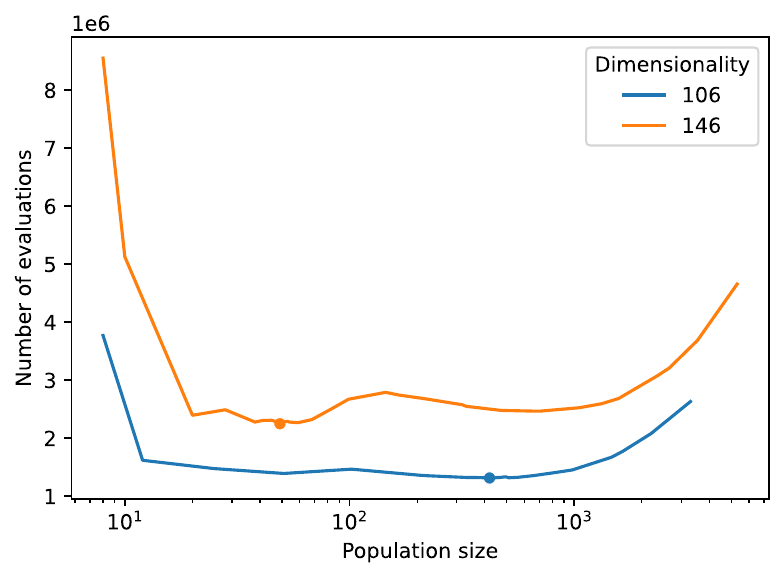}
    \caption{Population bisection results for Cigar show a wide basin of attraction. Dots indicate bisection result.}
    \label{fig:cigar_basin}
\end{figure*}

The population bisection results for iRV-GOMEA on the Cigar problem (and on various other problems as well) show a wide basin of attraction, which is illustrated for two problem dimensionalities in Figure~\ref{fig:cigar_basin}.
Cigar was omitted in the population sizing guideline as a large drop in best performing population size is observed between $\kappa=106$ and $\kappa=146$.

\subsection{Scaling Multiple Linkage Sets Results}
\begin{figure*}
    \centering
        \includegraphics[width=\textwidth]{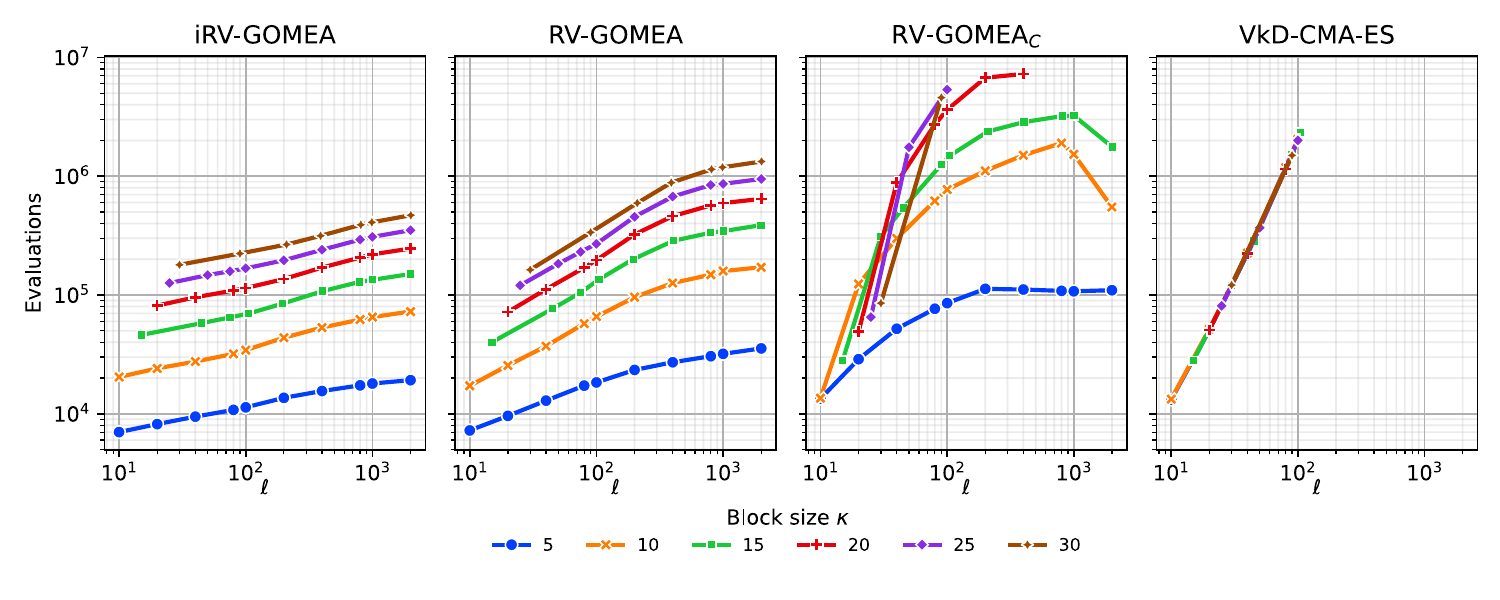}
    \caption{Required number of evaluations to reach VTR of the SoREB problem for different tested block sizes $\kappa$.}
    \label{fig:req_evals_gbo}
\end{figure*}

When scaling the number of FOS elements of size $\kappa$ in a GBO setting, the results are rather different than for single FOS elements in a BBO scenario.
Figure \ref{fig:req_evals_gbo} shows that as the problem dimensionality $\ell$ increases, when more blocks are present, iRV-GOMEA and RV-GOMEA show exceptional scalability by employing partial evaluations.
The smaller population sizes required for iRV-GOMEA to optimize the SoREB problem show a clear advantage over RV-GOMEA.
VkD-CMA-ES is incapable of partial evaluations and does not scale efficiently when $\ell$ increases.
Despite its decomposition capabilities, it requires the same number of evaluations for a problem size $\ell$ regardless of block size $\kappa$.
The CMA-ES sampling model in RV-GOMEA$_C$ shows better scalability due to partial evaluations.
Unexpected results are, however, observed for $\kappa \in [10, 15]$ and $\ell > 1000$, as those instances seem to require less evaluations than $\ell \in [400,800]$. 
For larger block sizes $\kappa > 20$ and problem dimensionalities $\ell > 100$, RV-GOMEA$_C$ can no longer find the optimum within the given budget, similar to VkD-CMA-ES.
The reason for these observations may well be that the CMA-ES sampling model and the GOM operator in RV-GOMEA${}_C$ are not seamlessly compatible. While RV-GOMEA${}_C$ upon its introduction was found to perform generally well~\cite{bouter_achieving_2021}, more extensive research in this particular combination was never performed. Our results indicate that this may be an interesting avenue of future research.

\subsection{Gray-Box Optimization Population Sizing Results}
\begin{figure*}
    \centering
        \includegraphics[width=\textwidth]{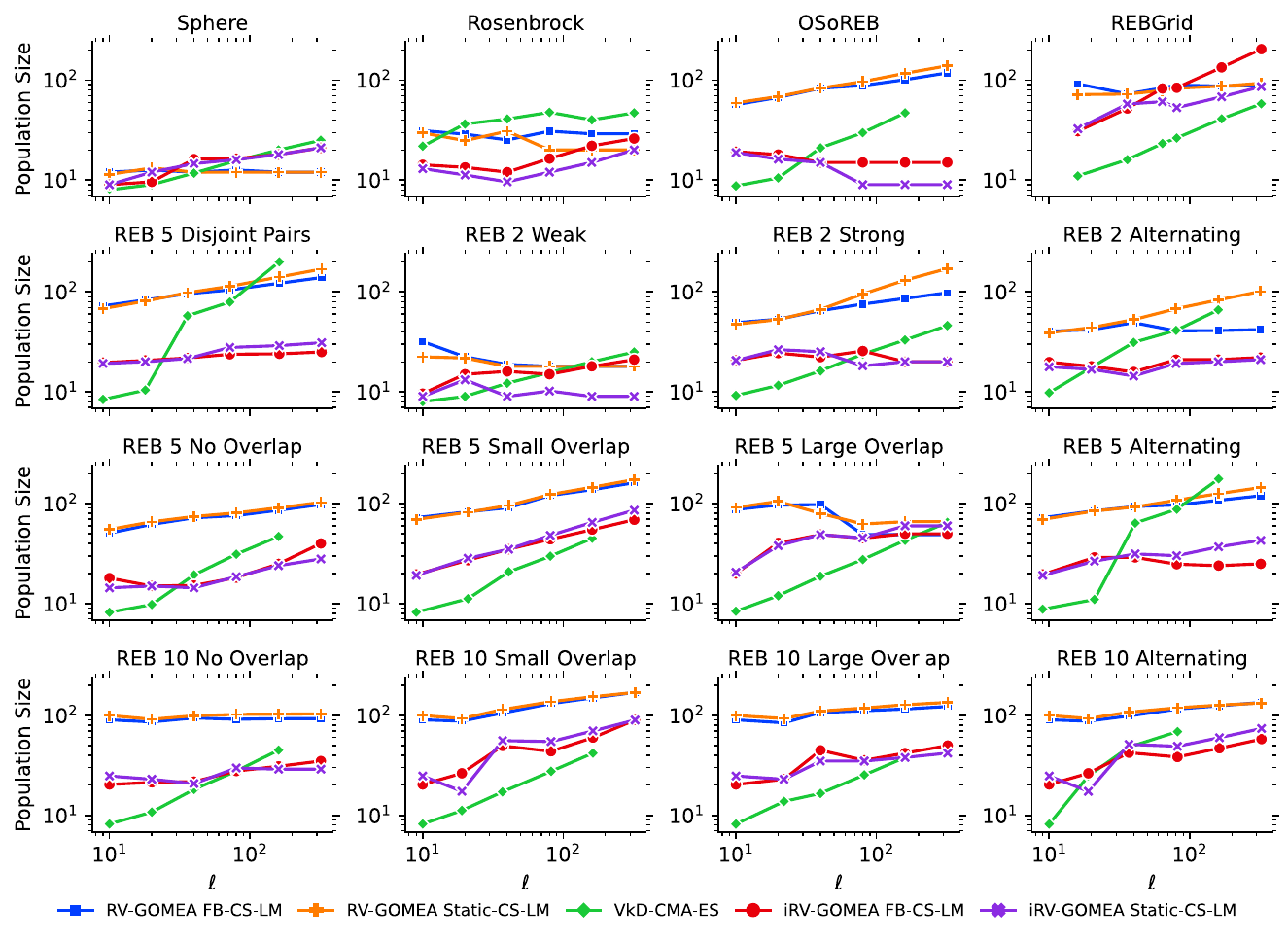}
    \caption{Optimized population sizes to reach VTR for all tested optimization problems.}
    \label{fig:req_popsize_gbo}
\end{figure*}

The bisected population sizes required to reach the optimum in as few evaluations as possible for the various conditional GBO problems are shown in Figure \ref{fig:req_popsize_gbo}.
For nearly all problems the results show a reduction in found population sizes for iRV-GOMEA vs RV-GOMEA, except for problems that can be optimized in a univariate manner.
VkD-CMA-ES requires smaller population sizes than RV-GOMEA variants especially for small problem dimensionalities, but on most problems, the population size for VkD-CMA-ES becomes larger than the minimum required by iRV-GOMEA as the problem dimensionality is increased.

\begin{figure*}
    \centering
        \includegraphics[width=\textwidth]{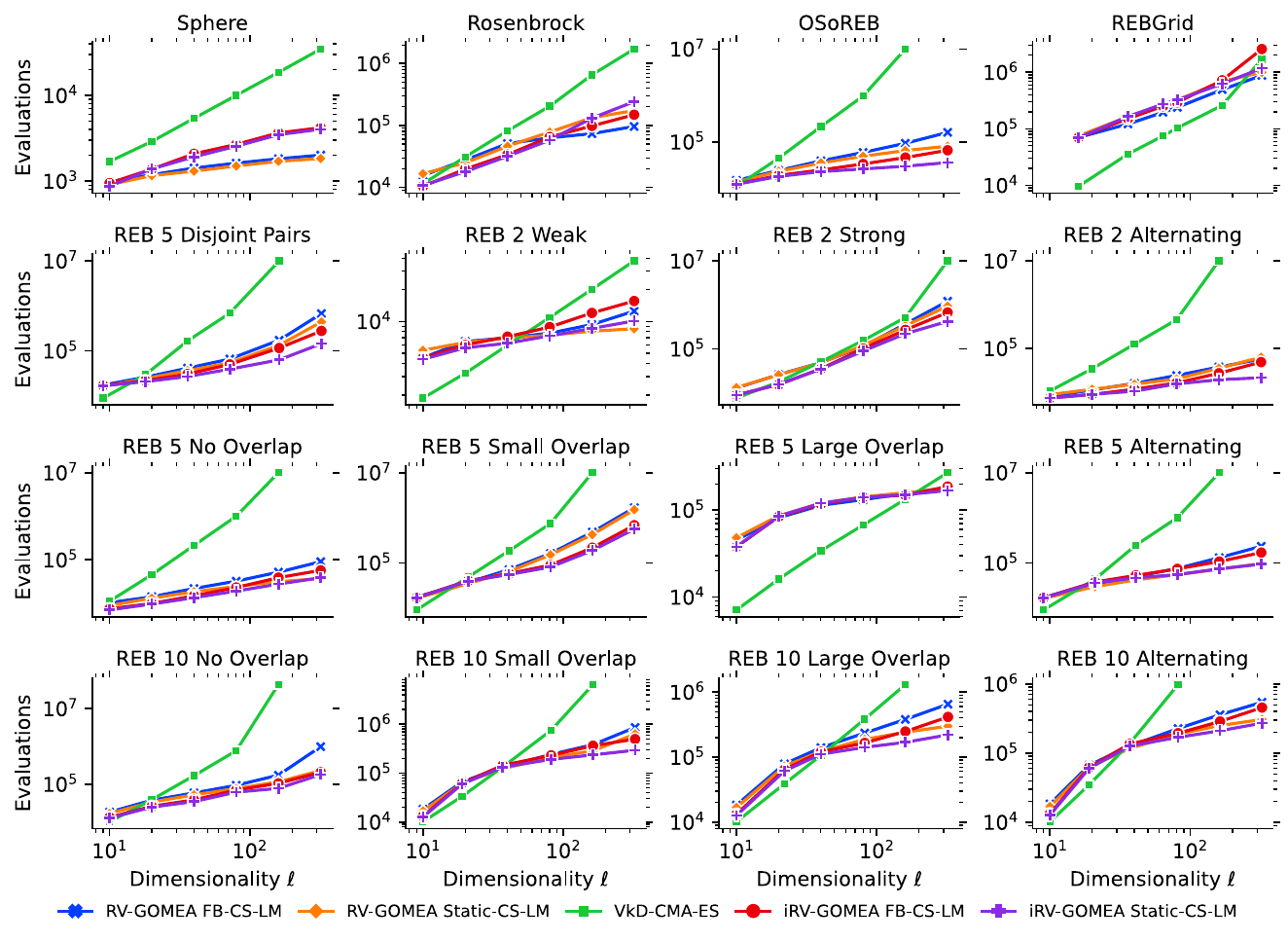}
    \caption{Evaluations to reach VTR for all tested optimization problems.}
    \label{fig:req_evals_gbo_cond}
\end{figure*}

Figure \ref{fig:req_evals_gbo_cond} shows the evaluations required to reach the VTR on all problems.
For fully separable problems like Sphere, iRV-GOMEA search shows deteriorating performance versus RV-GOMEA due to the efficiency of the GOM operator in univariate optimization in GBO.
For most other problem instances, iRV-GOMEA shows performance gains over RV-GOMEA.
iRV-GOMEA with the fitness-based linkage model is often on par or even outperforms the static linkage model based on the true VIG in RV-GOMEA.
Overall the advantage of the GOM operator in GBO is evident, as both iRV-GOMEA and RV-GOMEA show better scalability than VkD-CMA-ES when problem dimensionalities increase.
Similar to the non-conditional results, VkD-CMA-ES performs better than RV-GOMEA variants for smaller problem dimensions when optimization becomes more like a black box problem as there are few partial evaluations that can still be leveraged.


\end{document}